\documentclass[twocolumn,10pt]{asme2ej}

\usepackage{graphicx}
\usepackage{amsmath}
\usepackage{hyperref}   
\hypersetup{
	colorlinks=true,
	linkcolor=blue,
	citecolor=blue,
	urlcolor=blue,
}
\usepackage[square,numbers]{natbib}

\title{IN-PIPE ROBOT}

\author{Ravi Kant
}

\author{Gaurav Saha
}

\author{Arjun Kumar
}

\begin{document}

\maketitle    

\begin{abstract}
{This paper presents the arrangement of an in-pipe climbing robot that works using a clever differential part to explore complex associations of lines. Standard wheeled/continued in-pipe climbing robots are leaned to slip and take while exploring in pipe turns. The mechanism helps in achieving the first eventual outcome of clearing out slip and drag in the robot tracks during development. The proposed differential comprehends the down to earth limits of the standard two-yield differential, which is cultivated the underlying time for a differential with three outcomes. The mechanism definitively changes the track paces of the robot considering the powers applied on each track inside the line association, by clearing out the prerequisite for any unique control. The entertainment of the robot crossing in the line network in different bearings and in pipe-turns without slip shows the proposed arrangement's ampleness.  
}
\end{abstract}

\section{Introduction}

Pipe networks are inevitable, fundamentally used to transport liquids and gases in undertakings and metropolitan networks. Most often, the lines are covered to concur with the security rules and to avoid dangers. This makes audit and upkeep of lines genuinely testing. Covered lines are particularly disposed to deterring, utilization, scale game plan, and break initiation, achieving deliveries or damages that may incite grievous episodes. Different Inspection Robots \cite{han2016analysis} were proposed in the past to lead customary preventive assessments to avoid disasters. Divider crushed IPIRs with single, unique and crossbreed frameworks \cite{roslin2012review}, Pipe Inspection Gauges (PIGs) \cite{okamoto1999autonomous}, effectively controlled IPIRs with explained joints and differential drive units \cite{ryew2000pipe} were also generally thought of. In addition, bio-roused robots with crawler, inchworm, walking parts \cite{choi2007pipe}, and screw-drive\cite{kakogawa2013development} systems were in like manner showed to be fitting for different necessities. Regardless, most of them use dynamic controlling methods to guide and move inside the line. Dependence upon the robot's course inside the line added to the hardships, in like manner leaving the robot weak against slip in case traction control methods are not involved. The Theseus \cite{hirose1999design}, PipeTron \cite{debenest2014pipetron}, and PIRATE \cite{dertien2014design} robot series use separate segments for driving and driven modules that are interconnected by different linkage types. Each sections change or shift the bearing for organizing turns. Additionally, good unique controlling makes such robots reliant upon sensor data and profound estimation.

Pipe climbing robots with three adjusted modules are all the more consistent and give better movability. Earlier proposed detached line climbers \cite{vadapalli2019modular,suryavanshi2020omnidirectional} have utilized three driving tracks organized evenly to one another, like MRINSPECT series of robots \cite{roh2008modularized,roh2001actively,yang2014novel}. In such robots, to viably control the three tracks, their rates were pre-portrayed for the line turns. This addressed an imperative for the robot to orchestrate pipe-contorts exactly at a particular bearing contrasting with the pre-described rates \cite{vadapalli2019modular,suryavanshi2020omnidirectional,roh2008modularized,roh2001actively,yang2014novel}. In certifiable applications, the robot's course change accepting it experiences slip in the tracks during development. This limitation can be addressed by using a latently worked differential part to control the robot. MRINSPECT-VI \cite{kim2016novel,kim2013pipe} uses a multi-essential differential stuff part to control the paces of the three modules. In any case, for the division of the main thrust and speed to the three modules, the configuration of the differential is used. This philosophy made the chief yield (Z) to turn speedier than the other two outcomes (X and Y), making yield Z easily affected by slip \cite{kim2016novel}. This is caused considering the way that the consequences of the differential doesn't confer tantamount energy to the data. Other as of late proposed deals with serious consequences regarding differentials \cite{kota1997systematic,ospina2020sensorless} additionally followed a comparative design.

'Differential' kills the referred to limitation by recognizing indistinguishable outcome to enter dynamic relations \cite{vadapalli2021design}. In the imagined design all of the three outcomes are comparably affected by the information. This contributes for the robot to crash slip and drag toward any path of the robot during its development. In addition, the differential part in the line climber overhauls the comfort by diminishing the dependence on the powerful controls to travel through the line associations. The   system exactly moves power and speed from a lone commitment to the robot's three tracks through complex stuff trains, considering the stacks experienced by each track autonomously.
\section{In-pipe robot's design}

The CAD model of the proposed robot,  robot. A capricious nonagon central casing of the robot houses three modules circumferentially $120^\circ$ isolated from each other. The differential arranged inside the central instance of the robot drives the three tracks by its driving sprockets through. incline furnishes, that are related with yields. The organized point of view on the instrument and is clear in the resulting subsection. Each module houses a track and has openings for the four linkages to slide. The modules are pushed radially outwards with the help of direct springs mounted on the linkages (or shafts) i.e., the springs between the modules and the robot body. The fitting is a reference structure that limits the development of the modules past acceptable endpoints. Right when the robot is sent inside the line, the spring-stacked tracks go through detached evasion and presses against the line's internal dividers. It gives the fundamental traction to the robot to move. Each module in the robot can in like manner pack unevenly, point by point in fragment.

\vspace{-0.15in}
\begin{figure}[ht!]
\centering
\includegraphics[width=2.5in]{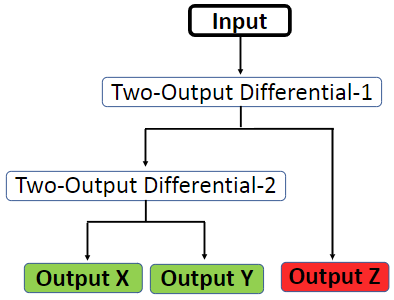}
\vspace{-0.15in}
\caption{\footnotesize Schematic}
\label{1}
\vspace{-0.15in}
\end{figure}
divider pressing part of the robot. The tracks, obliged on the modules, are pressed against the inner dividers of the line which give the significant balance to the robot to move. There is a driving sprocket that changes over the rotational development from the aftereffect of to translatory development for the robot. The divider crushing framework contains 4 direct linkages for all of the three modules. The modules have openings through which the straight linkages (or shafts) pass. They license the development of the modules in the winding headings from the turn of the robot.
\subsection{Mechanism}

The differential is the fundamental constituent of the proposed robot. The instrument contains a single data, three  differentials, three two-input open differentials and three outcomes. The differential's criticism is arranged at the central turn of the robot body. The three are coordinated uniformly around the data, with a place of $120{^\circ}$ between any two. The are fitted radially in the center. The single consequence of    construction the three outcomes of the  .Includes gear parts, for instance, ring gears, incline gears and side machine gear-piece wheels, while join ring gears, incline gears and side pinion wheels. The side machine gear-piece wheels ofis concurred with their close by side pinion wheels of   , to move the power and speed from     to it's adjoining   . The data  of the worm gear give development to     simuntaneously. Each two-yield differential then trade got development to its bordering two-input differentials , dependent upon the pile conditions experienced by its different side pinion wheels. The development got by the side pinion wheels of    further makes a translation of them to the three outcomes. The six differentials and work together to decipher development from the commitment to the three outcomes.

Right when   experience different weights, the side pinion wheels in    trades different weights to the side pinion wheels of    . Under this condition,     makes an understanding of differential speed to its adjoining   . Exactly when   experience comparable weight or no stack, all of the side pinion wheels experience a comparable weight making them turn at a comparative speed and power. As spread out, all of the outcomes share identical energy with the information. Besides, the outcomes in like manner share indistinct energy with each other. This results in the distinction in loads in one of the outcomes constraining a comparable effect on the other two outcomes that are undisturbed. The outcomes work with identical speeds when there is no pile or comparable weight circling back to all of the outcomes. The   part works its outcomes with differential speed when the outcomes are under changed weights. Exactly when one of the outcomes is working at a substitute speed while the other two outcomes are experiencing comparable weight, then, the two outcomes with comparable weights will work with identical speeds. The   part planned for the robot, pushes its the three tracks with indistinguishable speeds while moving inside a straight line. Regardless, while moving inside the line contorts, the differential changes the track speed of the robot with the ultimate objective that the track daring to all aspects of the more expanded distance turns faster than the track daring to all aspects of the more restricted distance. Allude to \cite{vadapalli2021modular,9635853} for additional figures, diagrams and data.
\section{Mathematical Modeling Of The Robt}

\subsection{Mathematical Modeling of the mechanism}

The kinematic plot shows the accessibility of the associations and joints of the   instrument. The differential's kinematic and dynamic conditions are arranged using the bond outline model. The data exact speed from the motor is likewise passed on to the three ring gear-tooth wheels of the two-yield differentials as. They turn at identical exact velocities and with comparable power for instance times is the stuff extent of the commitment to the ring gears. Moreover, a two-yield differential coordinates that the exact speed of its ring gear is the ordinary each season of the saucy paces of its two side pinion wheels. These two side gear-tooth wheels can turn at different rates while staying aware of identical power \cite{deur2010modeling}.
\begin{equation}
\omega=\frac{k(\omega _{01}+\omega _{02})}{2},
\label{1}
\end{equation}
From the bond outline model, we can infer that the three ring gears turns at identical exact paces and with comparable power for instance times the data dashing velocity, events the data power, is the stuff extent of the commitment to the ring gears. Also, in a two-yield differential, the exact speed of its ring gear is the ordinary all of the hour of the saucy paces of its two side pinion wheels. These two side machine gear-piece wheels can turn at different speeds while staying aware of identical power \cite{vadapalli2021design}.The daring rates are then implied their side machine gear-piece wheels. Side machine gear-piece wheels are related with the ultimate objective that they don't have any general development between the pinion wheels of a comparable pair, i.e., exact paces of side pinion wheels of a comparable pair is for the most part identical. Subbing in (\ref{1}). Basically, the outcome exact velocities. The outcome to side stuff association is contrasted with achieve the jaunty speed condition for the commitment to the outcomes. The rakish speeds of the ring pinion wheels to the side cog wheels, we achieve a connection among info and side cog wheels. Additionally, the result precise speeds are determined from the speeds of the individual ring gears to get a result to side stuff connection. Likening both the relations, we achieve the precise speed condition for the contribution to the results.

where is the exact speed of the data is the stuff extent of the ring gears to the outcome, while dapper paces of the outcome. The individual exact velocities of the side gear-tooth wheels. Consequently, the outcome exact paces are dependent upon the data dashing velocity ($\omega_u$) and the side stuff yields. Meanwhile, the powers of the ring gears is how much the powers of their looking at side pinion wheels. Like exact speed association for commitment to yields, by contrasting the ring pinion wheels with side pinion wheels association with the outcome to the ring gear association, an association between yield powers is gained.
\begin{equation}
\begin{split}
\tau_{O01}=\frac{k(\tau_u)}{3j}-\frac{(I_{01}\Dot{\omega}_{07}+I_{03}\Dot{\omega}_{08})}{j}, \ \hspace{0.25in}
\label{2}
\end{split}
\end{equation}
where is the exact speed of the data is the stuff extent of the ring gears  to the outcomes, while and are dashing rates of the outcome. Outputs are the individual exact velocities of the side gear-tooth wheels. Along these lines, the outcome exact paces are dependent upon the data jaunty speed and the side stuff yields. Meanwhile, the powers of the ring gears is how much the powers of their looking at side pinion wheels. Like exact speed association for commitment to yields, by contrasting the ring pinion wheels with side pinion wheels association with the outcome to the ring gear association, an association between yield powers to the data power is procured.
\begin{equation}
{\bf\omega _{O01}}=\frac{j(\omega _u)}{k}
\label{3}
\end{equation}
\begin{equation}
{\bf\tau _{O01}}=\frac{k(\tau_u)}{3j} - \frac{2(I_1\dot{\omega}_1)}{j}
\label{4}
\end{equation}
Conditions \eqref{3} and \eqref{4} presents the interest of the differential to give comparable development ascribes in all of the three outcomes that experiences identical weights or when left unconstrained. Notwithstanding, when the stuff parts experience an obstacle across a crossing point, the dashing rate and the power changes depending upon the external resistive power. The outcome speed of the differential outcomes are comparable to the speed of the driving module sprockets.Therefore, the outcome speeds ${\omega_{O1}}$, ${\omega_{O2}}$ and ${\omega_{O3}}$ of the differential are changed over into track speeds $v_{tA}$, $v_{tB}$ and $v_{tC}$. The data speed for the robot is 120 rpm, thusly making an understanding of 12 rpm to the outcomes under comparable stacking conditions. The sprocket broadness is consistent ($D_s$= 80 mm) for all of the three tracks. Ho Moon Kim et al. \cite{kim2013pipe}, in their paper suggested a strategy for working out the specific paces of the three tracks inside pipe turns. Tolerating that the robot enters the pipeline with the plan showed is the mark of the line turn, R is the breadth of rhythmic movement of the line bend and r is the scope of the line. The speed of the track is gotten from
\begin{equation}
v_{A} = v(\frac{R-r\cos({\mu})}{R})
\label{eqn01}
\end{equation}
Similarly, the paces $v_{tB}$ and $v_{tC}$ for their singular tracks B and C are gotten. The robot is inserted at different heading of the modules concerning OD. The changed speeds for the not set in stone for wind pipes in headings $\mu$ = $0^\circ$, $\mu$ = $30^\circ$, $\mu$ = $60^\circ$.

\subsection{Compression}

The straight springs in the module gives robot the flexibility to organize turns with practically no issue. The best strain possible in each module is $16 mm$. There are additional versatilities in the module openings, with the objective that upside down pressure is possible. This helps the robot with overcoming obstructions and oddities in the line network that it may examine genuine applications. The front completion of the module is compacted absolutely however the posterior is in its most noteworthy widened state possible inside the line. The most outrageous disproportionate tension allowed in a lone module of the robot. Thusly, $\phi$ is the best point the module can pack unevenly. Allude to \cite{vadapalli2021modular,9635853} for additional figures, diagrams and data.
\vspace{-0.1in}
\begin{figure}[ht!]
\vspace{-0.1in}
\centering
\includegraphics[width=0.75in]{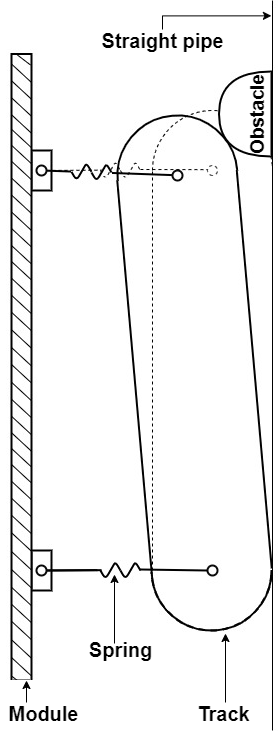}
\vspace{-0.12in}
\caption{\footnotesize Module compression}
\label{asym(1)}
\vspace{-0.1in}
\end{figure}

\section{Simulations of the Robot}

Diversions were directed to examine and support the development limits of the robot in various line associations. A comparable will give us more encounters into the components and direct of the made  robot in real testing conditions. Hereafter, multi-body dynamic propagations was acted in MSC Adams by changing over the arrangement limits into an enhanced reenactment model. To lessen the amount of moving parts in the model and to decrease the computational weight, the tracks were modified into roller wheels. Each module houses three roller wheels in the enhanced model. Thusly, the contact fix given by the tracks to the line dividers are diminished from 10 contact regions to 3 contact regions for each module. The reenactment limits like the track speeds ($v_{tA}$, $v_{tB}$, and $v_{tC}$) and the module strain for each track A,B and C were analyzed. Multiplications were performed by implanting the robot in three unmistakable headings of the module ($\mu$ = 0${^\circ}$, $\mu$ = 30${^\circ}$, $\mu$ = 60${^\circ}$) in both the straight lines and line turns. The robot is repeated inside a line network arranged by ASME B16.9 standard NPS 11 and plan 40. The diversions were coordinated for four examination circumstances in the line network containing Vertical region, Elbow section (90${^\circ}$ bend), Horizontal portion and the U-fragment (180${^\circ}$) for different bearings ($\mu$ = 0${^\circ}$, $\mu$ = 30${^\circ}$, $\mu$ = 60${^\circ}$) of the robot. The outright distance of the line structure is $D_{pipe}$ = 3,023.49mm. The distance went by the robot ($D_{R}$) in not set in stone from point of convergence of the robot body and the track's solitary distance still up in the air from the center roller wheel mounted in each module. Along these lines, we get the outright robot's way, by removing the robot's length from $D_{pipe}$ (i.e., $D_{R}$ =$D_{pipe}$ - $(L_R)$ = 2,823.49mm, where $L_R$= 200mm).The robot's way in vertical climbing and the last level region is assessed by deducting $L_R/2$ from their singular portion length. The information ($U$) of the   is given a reliable dapper speed of 120rpm ($\omega_u$= 120rpm) and development of the robot including the track speeds ($v_{tA}$, $v_{tB}$, and $v_{tC}$) are considered in the reenactment.

In the vertical fragment and level section, the robot follows a straight way. Thus, the tracks experience identical weights on all of the three modules in both the investigations. Thusly, the differential gives identical velocities to all of the three tracks, similar to the robot's ordinary speed $v_R$. The saw track speeds in the diversion for the heading $\mu$ = 0$^\circ$, is $v_R$= $v_{tA}$= $v_{tB}$= $v_{tC}$= 50.03 mm/sec. In like manner, for $\mu$ = 30$^\circ$, the rates are $v_R$= $v_{tA}$= $v_{tB}$= $v_{tC}$= 50.22 mm/sec and for $\mu$ = 60$^\circ$, the velocities are $v_R$= $v_{tA}$= $v_{tB}$= $v_{tC}$= 51.36 mm/sec. Thusly, all of the characteristics contrast with the theoretical results with a level out rate botch (APE) lesser than 2.2\%. This error assesses the veritable proportion of deviation from the speculative worth \cite{armstrong1992error}. To organize a fundamental length of 550 mm ($D_{R}$ = 550 - $(L_R/2)$= 450 mm) in vertical climbing, the robot expects 0 to 9 seconds. the robot's ability to climb the line up against gravity. Starting from 24 seconds till 31 seconds, the robot moves a distance of 350 mm in the chief even region. In the more humble level fragment of distance 150mm ($D_{R}$ = 150 - $(L_R/2)$= 50 mm), the robot moves from 59 to 60 seconds.

In the elbow region ($90^\circ$ curve) and U-fragment ($180^\circ$ bend), the robot moves at a consistent distance to the point of convergence of curve of the line. The   framework changes the outcome speeds of the tracks $v_{tA}$, $v_{tB}$ and $v_{tC}$ as demonstrated by the partition from the point of convergence of state of the line. In every one of the three headings of the robot, the outer module tracks goes faster to travel a more broadened distance, while the internal module tracks turns all the more delayed to dare to all aspects of the most restricted distance than the range of bend of the line curve. In pipe contorts, the entertainment velocities of each track is shown up at the midpoint of seperately to harsh the saw track speeds without changes. The approximated velocities of each track is then stood out from their specific speculative rates with track down the by and large rate botch (APE). For the course $\mu$ = 0$^\circ$, the outside modules (B and C) move at a typical speed of (58.7 mm/sec and 57.8 mm/sec), while the internal module (A) move at an ordinary speed of 33.62mm/sec. These characteristics connect with the theoretical characteristics $v_{tB}$ = $v_{tC}$ = 58.51 mm/sec and $v_{tA}$ = 33.69mm/sec with an APE of 1.2\%. Moreover, the track speeds $v_{tA}$, $v_{tB}$ and $v_{tC}$ for $\mu$ = 30$^\circ$, organizes with the typical worth of the proliferation results ($v_{tC}$= 50.3 mm/sec, $v_{tB}$= 63.8 mm/sec, $v_{tA}$= 37.3 mm/sec) with an APE of 3.8\%. Likewise, the track speed a motivator for $\mu$ = 60$^\circ$, contrast with the proliferation results ($v_{tB}$= 68.5 mm/sec, $v_{tA}$= 40.2 mm/sec and $v_{tC}$= 41.3 mm/sec) with an APE of 2.5\%. Toward each path of both the straight and turn regions, the slip-up regard is incredibly unimportant and they can be credited as a result of outside factors in obvious conditions like disintegration. Thusly, irrelevant speed changes occurs in the proliferation plot. From 9 to 24 seconds, the robot organizes the elbow region (90$^\circ$ bit) of distance 657.83 mm, while it requires 31 to 59 seconds to travel 1315.66 mm in the U-portion (180$^\circ$ bend). The robot's ability to explore in pipe-turns. for $\mu$ = 0$^\circ$ shows that the outside modules (B and C) move at the speed extent of (52-60mm/sec) while the internal module (A) move in a speed extent of (30-37.25mm/sec). These characteristics exist in the speculative characteristics $v_{tB}$ = $v_{tC}$ = 58.51 mm/sec and $v_{tA}$ = 33.69mm/sec. The track speeds $v_{tA}$, $v_{tB}$ and $v_{tC}$ for $\mu$ = 30$^\circ$, organizes with the reenactment results ($v_{tC}$= 49-52mm/sec, $v_{tB}$= 60-75mm/sec, $v_{tA}$= 32.5-40mm/sec). Additionally, the track speeds a motivator for $\mu$ = 60$^\circ$, exist in the reenactment results($v_{tB}$= 62-75mm/sec and $v_{tA}$= $v_{tC}$= 32.5-48). The speed range referred to from the diversion is taken from the base to the most outrageous apexes of rates for each plots. Allude to \cite{vadapalli2021modular,9635853} for additional figures, diagrams and data.

The proliferation results for the track speeds $v_{tA}$, $v_{tB}$, $v_{tC}$ and the robot $v_{R}$ in different headings ($\mu$ = 0$^\circ$, $\mu$ = 30$^\circ$, $\mu$ = 60$^\circ$), organizes with the theoretical results got in region III. It is seen from the reenactment that in 60 seconds, the robot explores all through the line network reliably at the inserted bearing. The result connect with the speculative calculation ($D_{R}$/$v$= 3016.49/50.24 = 60.04 sec). This endorses that the differential discards slip and drag in the tracks of the line climber every which way with no development adversities. In the propagation, the robot is seen with no slip and drag every which way, which further impacts in diminished tension effect on the robot and extended development flawlessness. \textbf{\small Radial flexibility:\normalsize} The track in the modules catch to the inward mass of the line to give balance during development. The springs are at first pre-stacked by a tension of 1.25 mm in every one of the three modules likewise when inserted in the vertical line region. In straight lines, the robot moves at the basic pre-stacked spring length. The disfigurement length increases by 1.5 mm for the inside and the outside modules when the robot is moving near elbow region and U-section. This deformation explains the extended flexibility thought about the modules to travel through the developing cross-portion of the line broadness in the bends during development.

\section{Conclusion}
The  robot robot is given the smart differential to control the robot unequivocally with no unique controls. The differential has an indistinguishable outcome to incorporate energy, whose show is absolutely like the convenience of the standard two outcome differential. The generation results support powerful intersection of staggering line networks with bends of up to 180$^\circ$ in different headings without slip. Taking on the differential part in the robot achieves the sharp delayed consequence of clearing out the slip and drag every which way of the robot during the development. At the present, we are encouraging a model to perform researches the proposed arrangement.
%
% ...

\bibliographystyle{asmems4}

\bibliography{asme2e}

\end{document}